\documentclass[10pt,twocolumn,letterpaper]{article}

\usepackage[pagenumbers]{iccv} 

\definecolor{iccvblue}{rgb}{0.21,0.49,0.74}
\usepackage[pagebackref,breaklinks,colorlinks,allcolors=iccvblue]{hyperref}
\usepackage{marvosym}

\newcommand{\figref}[1]{Fig.~\ref{#1}}

\newcommand{\secref}[1]{Sec.~\ref{#1}}
\newcommand\blfootnote[1]{%
\begingroup
\renewcommand\thefootnote{}\footnote{#1}%
\addtocounter{footnote}{-1}%
\endgroup
}

\title{CoGen: 3D Consistent Video Generation via Adaptive Conditioning for Autonomous Driving}

\author{Yishen Ji\textsuperscript{1,2} \,\, 
Ziyue Zhu\textsuperscript{2,3} \,\,
Zhenxin Zhu\textsuperscript{2} \,\, 
Kaixin Xiong\textsuperscript{2} \,\,
Ming Lu\textsuperscript{4}  \,\,
Zhiqi Li\textsuperscript{1} \\
Lijun Zhou\textsuperscript{2,$\dagger$}  \,\,
Haiyang Sun\textsuperscript{2,$\dagger$}  \,\,
Bing Wang\textsuperscript{2,\Letter} \,\,
Tong Lu\textsuperscript{1,\Letter} \\ 
\\
\textsuperscript{1}Nanjing University \hspace{1em} \textsuperscript{2}Xiaomi EV \hspace{1em} \textsuperscript{3}Nankai University\hspace{1em} \textsuperscript{4}Peking University \\
{\tt\small jiyishen929@smail.nju.edu.cn}
}
\begin{document}
\maketitle

\blfootnote{
\textsuperscript{$\dagger$}project leader. \textsuperscript{\Letter}corresponding author.
}

\begin{abstract}
Recent progress in driving video generation has shown significant potential for enhancing self-driving systems by providing scalable and controllable training data. 
Although pretrained state-of-the-art generation models, 
guided by 2D layout conditions (\eg, HD maps and bounding boxes), 
can produce photorealistic driving videos, 
achieving controllable multi-view videos with high 3D consistency remains a major challenge. 
To tackle this, 
we introduce a novel spatial adaptive generation framework, 
\textbf{CoGen}, 
which leverages advances in 3D generation to improve performance in two key aspects: 
\(\left( i \right)\) To ensure 3D consistency, 
we first generate high-quality, 
controllable 3D conditions that capture the geometry of driving scenes. 
By replacing coarse 2D conditions with these fine-grained 3D representations, 
our approach significantly enhances the spatial consistency of the generated videos.
\(\left( ii \right)\) Additionally, 
we introduce a consistency adapter module to strengthen the robustness of the model to multi-condition control.
The results demonstrate that this method excels in preserving geometric fidelity and visual realism, 
offering a reliable video generation solution for autonomous driving. 
The project for CoGen is available at: \href{https://xiaomi-research.github.io/cogen/}{https://xiaomi-research.github.io/cogen/}.
\end{abstract} 
\section{Introduction}
\label{sec:introduction}

\begin{figure*}[ht]
    \centering
    \includegraphics[width=0.95\linewidth]{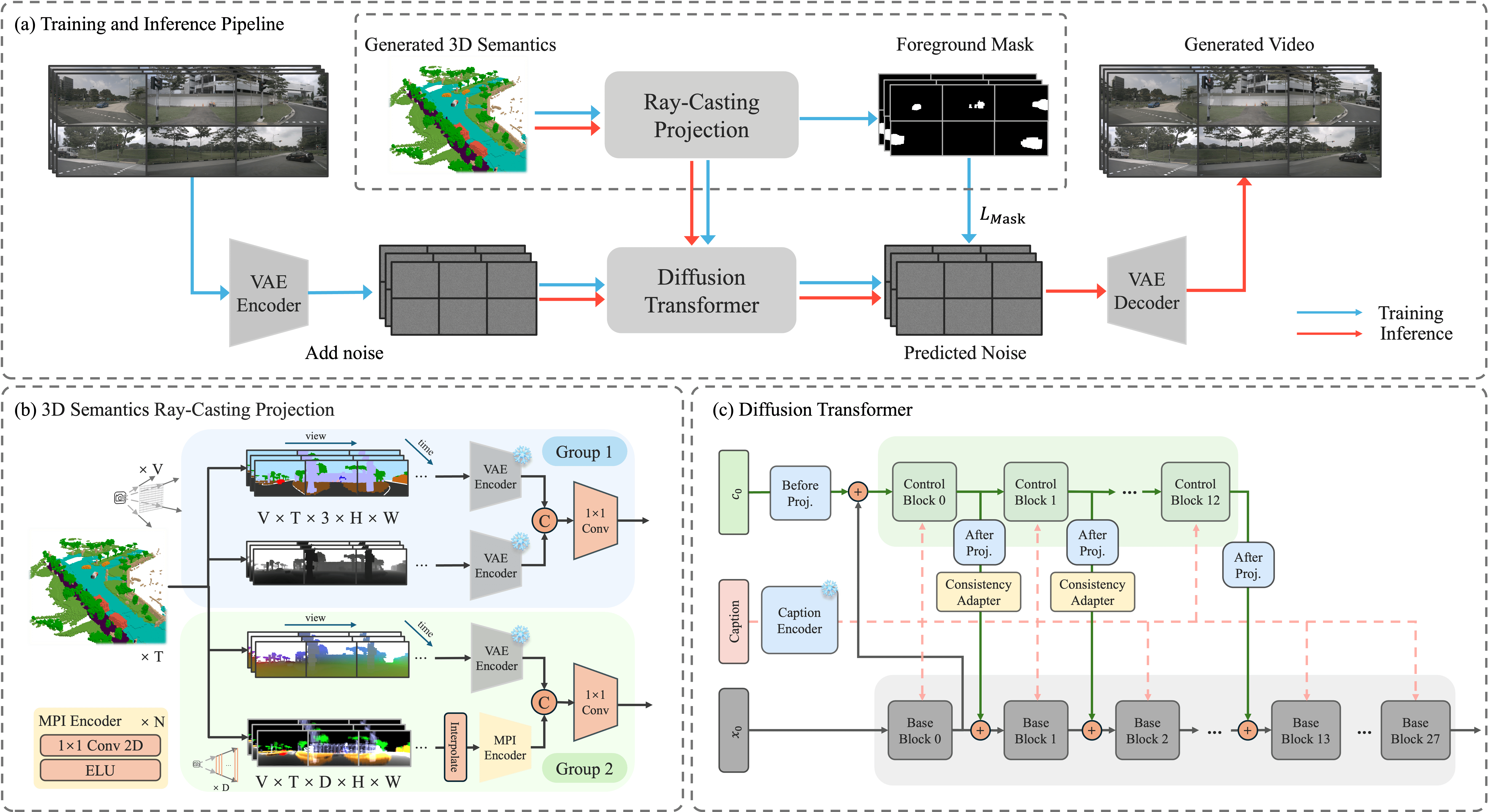}
    \caption{Overview of our model. (a) Training and inference pipeline. Using BEV maps as conditions, we generate temporal 3D semantics sequences, which are then projected and encoded to provide guidance for video generation. During projection, a foreground object mask is created and incorporated into training with a foreground mask loss reweight, enhancing supervision for foreground generation quality. (b) Details of 3D semantics projection and encoding. Various forms of guidance are fused through $1\times1$ convolutions. (c) Illustration of our diffusion transformer architecture.}
    \label{fig:model}
\end{figure*}

Developing a safe and reliable autonomous driving system requires diverse, non-conventional data to effectively train robust models. However, acquiring such data is often prohibitively challenging and costly. To address this, high-quality synthesis of driving scenes~\cite{gao2023magicdrive, jiang2024dive, gao2024magicdrivedit, wang2024drivedreamer, BEVGen, li2024drivingdiffusion, wang2024drivingintothefuture, li2024uniscene,chen2024unimlvg} has emerged as an increasingly pivotal solution in autonomous driving, offering a highly effective means to significantly reduce the costs of gathering and labeling real-world data. These approaches leverage advancements in generative models~\cite{LDM, dit, vdm, OpenSora-VAE-v1.2} to generate highly photorealistic images and videos, thereby enabling robust training and comprehensive evaluation for a wide range of downstream tasks~\cite{li2024bevformer, wang2021fcos3d, huang2021bevdet, jiang2023vad, hu2023planning,li2022coda}.

Despite significant advancements in generative models, synthesizing multi-view driving sensor data remains a significant challenge. Unlike general image/video generation methods~\cite{OpenSora-VAE-v1.2, dit, vdm,yang2024cogvideox}, which focus on producing single-view outputs that appear realistic to human observers, multi-view driving data must satisfy two critical requirements: photorealism and strict 3D consistency. The latter ensures geometric coherence across camera viewpoints, a necessity for reliable downstream tasks in autonomous driving systems.

Current approaches~\cite{jiang2024dive, gao2024magicdrivedit, gao2023magicdrive} rely on cross-attention modules and street layouts to maintain cross-view consistency,
however, achieving robust 3D consistency in the generated videos remains an ongoing challenge.
Meanwhile, advanced video generation methods for driving typically rely on low-fidelity geometric conditions,
such as 2D maps or bounding boxes,
overlooking 3D details and limiting the realism of the generated videos.
Further, recent studies~\cite{jiang2024dive, gao2023magicdrive} have shown that multi-condition control outperforms single-condition approaches.
Devising robust strategies to seamlessly integrate diverse conditions into these models is critical for enhancing performance.

To address these challenges,
we introduce \textbf{CoGen},
a novel spatially-adaptive framework for driving video generation.
To enhance spatial consistency, we avoid directly using 2D maps and bounding boxes as conditions for video generation.
Instead, we leverage them to construct a temporal 3D generative model that generates the 3D semantics squences of driving scenes.
Building on this foundation,
we propose four complementary 3D-semantics-informed conditions for video generation: Semantic Map, Depth Map, Coordinate Map, and Multi-Plane Image (MPI)~\cite{tucker2020single_mpi, szeliski1999stereo_mpi, zhou2018stereo}.
Derived via ray casting from 3D semantic representations,
these projections surpass traditional 2D conditions by more effectively capturing geometric structure and semantic context.
By preserving temporal 3D fidelity and multi-depth details,
they enable precise modeling of dynamic elements (\eg, vehicles and pedestrians) across multiple views.
Moreover, to enable our video generation models to adapt to multiple conditions,
we design a lightweight Consistency Adapter.
Positioned between the control blocks and the backbone network,
the adapter employs depthwise separable convolutions, 3D convolutions, and multi-head self-attention
to ensure smooth inter-frame transitions and long-range coherence.
This architecture allows the model to efficiently incorporate additional conditions via post-training,
without updating the parameters of the ControlNet or the backbone diffusion model.

Experimental results on the nuScenes dataset demonstrate that our approach achieves state-of-the-art video generation quality, as measured by metrics such as Fréchet Video Distance (FVD), while preserving high visual fidelity. This work offers a robust, controllable framework for synthesizing high-quality driving scene videos, enhancing the capabilities of autonomous driving simulation.
Our contributions can be summarized as follows:
\begin{itemize}
    \item We systematically investigate four 3D-semantics-derived guidance projections, validating their ability to enhance geometric fidelity and visual realism in video generation.
    \item We introduce a Consistency Adapter to improve the model's adaptability to multiple conditions, significantly enhancing motion coherence across frames.
    \item Experiments on benchmark datasets and metrics demonstrate that our approach achieves state-of-the-art performance in driving video generation.
\end{itemize}

\section{Related Works}
\label{sec:related_works}

\paragraph{Driving Scene Video Generation.}
High-quality video generation is a crucial method for generating autonomous driving training data. With the advancement of diffusion models\cite{zhang2023adding,LDM,dit,vdm,yang2024cogvideox,peng2024controlnext}, diffusion-based video generation methods have gained attention. Methods such as BEVGen\cite{BEVGen}, MagicDrive\cite{gao2023magicdrive}, and Panacea\cite{wen2024panacea} have explored image generation using HD maps, bounding boxes, and other control conditions. Since video data are more commonly used form in autonomous driving training data, methods like DriveDreamer\cite{wang2024drivedreamer}, SubjectDrive\cite{huang2024subjectdrive}, and UniMLVG\cite{chen2024unimlvg} have investigated the controlled generation of multi-frame, multi-view videos. However, the aforementioned condition controls based on HD maps and boxes suffer from geometric simplification and projection inconsistency issues. 
Some works, such as UniScene~\cite{li2024uniscene}, DrivingSphere~\cite{yan2024drivingsphere}, and InfinityCube~\cite{lu2024infinicube}, use occupancy as an intermediate representation.
Despite these improvements, existing methods still face challenges in generating high-fidelity depth-aware scene structures, especially for occluded and distant objects.
Our method builds upon the use of 3D semantics as a condition and further explores various forms of 3D semantics as guidance, achieving state-of-the-art video generation results.

\paragraph{Conditional Generation}
Recent advancements in driving scene video synthesis rely on a variety of control conditions to guide the generation process, with high-definition (HD) maps and instance bounding boxes being widely adopted for their ability to provide precise environmental structuring and object localization. 
These conditions can be categorized as follows: (1) Layout conditions, including bird's-eye-view (BEV) layouts~\cite{wen2024panacea,huang2024subjectdrive,ma2024delphi,jiang2024dive} for spatial consistency and 3D layouts~\cite{li2024drivingdiffusion,wang2024driving_drivewm} for geometric cues; Action inputs~\cite{wang2024driving_drivewm} for dynamic synthesis.
While these methods have improved motion control and foundational geometric modeling, they face two critical limitations: (1) \textbf{Geometric Simplification}: reliance on 2D planar representations (\eg, BEV layouts)~\cite{yang2023bevcontrol} and object-level constraints (\eg, bounding boxes) fails to capture intricate 3D structural details, such as building facades and road geometries, resulting in static scenes with limited spatial diversity~\cite{gao2023magicdrive,wen2024panacea}; (2) \textbf{Projection Inconsistency}: the 3D-to-2D projection process\cite{li2024drivingdiffusion} struggles to maintain spatial relationships in occlusion-heavy scenarios, leading to visual artifacts like floating vehicles and misaligned objects. These challenges degrade the visual realism and functional utility of generated videos, underscoring the need for robust 3D representations. 
Our method addresses these limitations by leveraging 3D semantics projections to preserve 3D structural fidelity and enhance spatiotemporal coherence, as detailed in subsequent sections.

\section{Methods}
\label{sec:methods}

In this section, we first revisit the concept of conditional diffusion models as a preliminary (\ref{sec: pre}).
Then we elaborate on the technical details of our framework including:
our fine-grained 3D conditions generator for ensuring spatial consistency (\ref{sec: guidance_projections}),
a 3D geometry-aware diffusion transformer architecture for video generation (\ref{sec: video_generation}),
and a lightweight adapter for adapting to multi-grained conditions (\ref{sec: consistency_adapter}).
Finally, we introduce how our loss design enhances the generation of foreground details (\ref{sec: mask_loss}).
An overview of our framework is presented in \figref{fig:model}.

\subsection{Preliminaries: Conditional Diffusion Models}
\label{sec: pre}

Latent Diffusion Models (LDMs)~\cite{LDM} address the high computational cost of diffusion models by operating in a lower-dimensional latent space. Given an image $\mathbf{x}$, an encoder $E$ is used to obtain the corresponding latent representation $\mathbf{z}=E(\mathbf{x})$. The forward diffusion process in the latent space is defined as a gradual noising process:
\begin{equation}
    q(\mathbf{z}_t|\mathbf{z}_0) = \mathcal{N}\Bigl(\mathbf{z}_t; \sqrt{\bar{\alpha}_t}\,\mathbf{z}_0, (1-\bar{\alpha}_t)\mathbf{I}\Bigr),
\end{equation}
where $\bar{\alpha}_t = \prod_{i=1}^{t}(1-\beta_i)$ with $\beta_i$ being a variance schedule. The reverse process is modeled by a neural network $\epsilon_\theta$ and parameterized as:
\begin{equation}
    p_\theta(\mathbf{z}_{t-1}|\mathbf{z}_t) = \mathcal{N}\Bigl(\mathbf{z}_{t-1}; \mu_\theta(\mathbf{z}_t,t), \Sigma_\theta(\mathbf{z}_t,t)\Bigr).
\end{equation}
Finally, a decoder $D$ maps the denoised latent variable back to the image space, i.e., $\mathbf{x}'=D(\mathbf{z}_0)$.

ControlNets~\cite{controlnet2023} extend LDMs by incorporating additional conditioning signals $\mathbf{c}$ to provide finer control over the generation process. In particular, the reverse diffusion process is modified to condition on $\mathbf{c}$:
\begin{equation}
    p_\theta(\mathbf{z}_{t-1}|\mathbf{z}_t,\mathbf{c}) = \mathcal{N}\Bigl(\mathbf{z}_{t-1}; \mu_\theta(\mathbf{z}_t,t,\mathbf{c}), \Sigma_\theta(\mathbf{z}_t,t,\mathbf{c})\Bigr).
\end{equation}
The conditioning variable $\mathbf{c}$ can encapsulate various forms of guidance such as spatial maps, semantic layouts, or other task-specific signals. By integrating $\mathbf{c}$, ControlNets offer enhanced control, enabling more precise manipulation of the generation and improved quality of synthesized images.

Overall, the combination of latent diffusion models and ControlNets forms a powerful framework for generating high-quality images with explicit controllability.
Recent driving video generation methods~\cite{gao2023magicdrive, jiang2024dive, gao2024magicdrivedit} typically employ diffusion models for video synthesis. 
In these approaches, the generation process is conditioned on projected 2D layout cues (\eg, HD maps and bounding boxes) that are encoded by ControlNets, 
enabling control over the synthesized content to some extent.

\subsection{Temporal 3D Semantics Conditions Generator}
\label{sec: guidance_projections}

\begin{figure}[ht]
    \centering
    \includegraphics[width=0.95\linewidth]{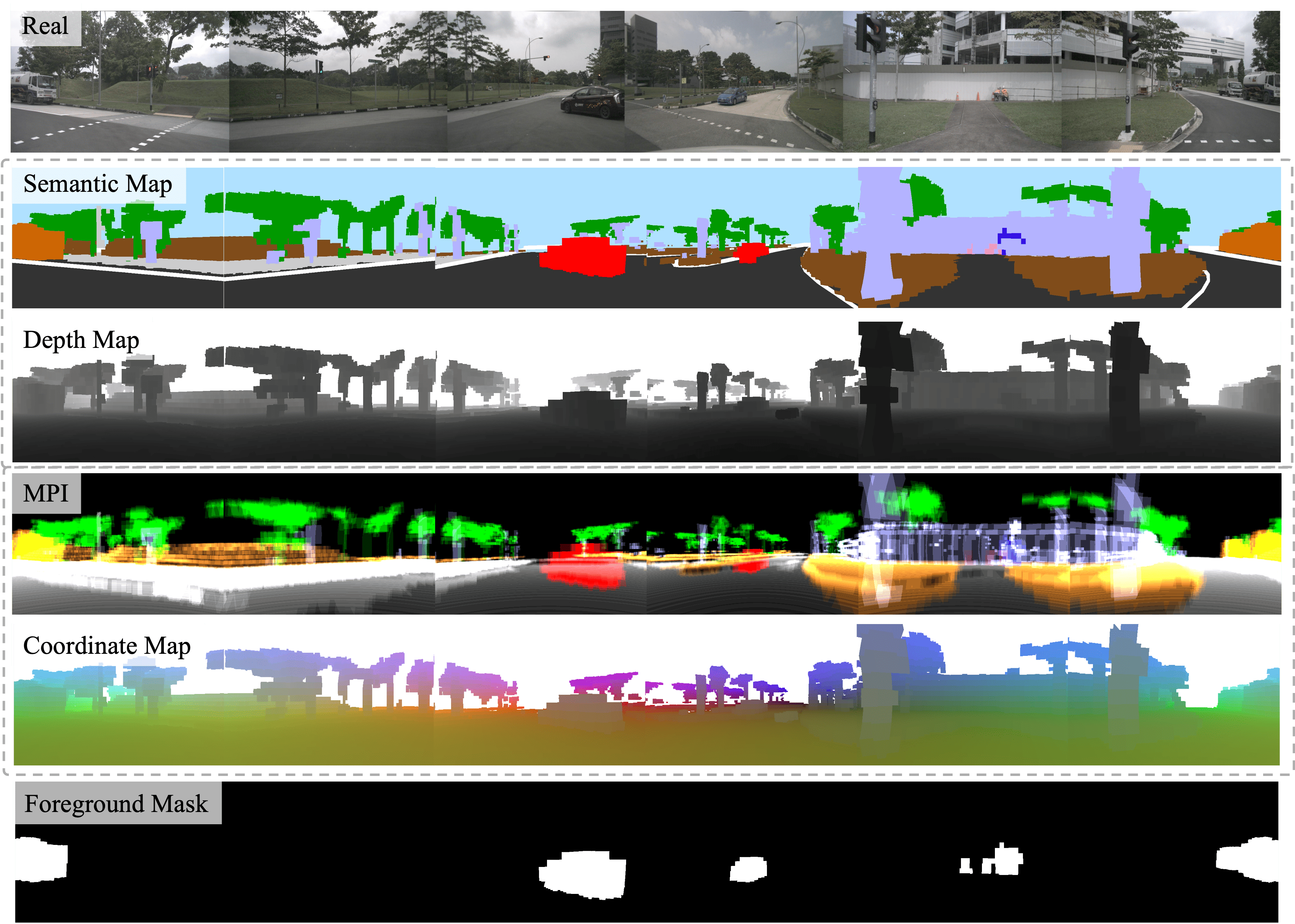}
    \caption{Visualization of the 3d semantics conditions used for video generation. Each condition is derived by projecting the 3D semantics grid into the camera view using ray casting, capturing essential geometric and semantic information for enhanced video generation.}
    \label{fig:condition}
\end{figure}

Although recent driving video generation methods~\cite{jiang2024dive, gao2023magicdrive, wang2024drivedreamer} use 2D maps and bounding boxes for conditioning, 
they often lack sufficient 3D consistency for downstream tasks. 
To improve spatial coherence, 
we propose a temporal 3D semantics generator that produces diverse, fine-grained 3D layouts. 
These ``Generated 3D Semantics'' (see \figref{fig:model}) are integrated as additional conditioning signals in the video generation process. 
Specifically, following~\cite{tong2023occnet}, 
the 3D semantics \( \mathbf{S} \in \mathbb{R}^{H \times W \times D} \) discretize the ego vehicle's surroundings into a voxel grid where each voxel is assigned a semantic label. 
We project the generated 3D semantics onto image plane as generation condition.
Compared to 2D conditions, 
the 3D semantics provide both detailed semantic cues and precise geometric guidance.

Drawing inspiration from 3D semantics generation methods~\cite{wang2024occsora, ren2024xcube, gu2024dome}, 
we employ a 3D Variational Autoencoder (VAE) 
as a tokenizer and a diffusion transformer to train the generator.   
To ensure temporal consistency across the conditions, 
our VAE encoder 
compresses temporal 3D semantics into a single latent representation 
\( \mathbf{z} \in \mathbb{R}^{f \times c \times h \times w} \), which is then used to train the diffusion transformer. 
The VAE decoder is responsible for reconstructing 3D semantics from tokens.
The loss function for training the diffusion model is defined as:
\begin{equation}
    \mathcal{L}_{\text{sem}} = \mathcal{L}_{\text{CE}}(\mathbf{S}, \hat{\mathbf{S}}) + \alpha \mathcal{L}_{kl} + \beta \mathcal{L}_{\text{lovasz}}(\mathbf{S}, \hat{\mathbf{S}}),
\end{equation}
where cross entropy, KL divergence, and Lovasz-softmax losses are employed.
Then, 2D maps and object bounding boxes serve as control conditions to train a light-weight diffusion transformer for temporal 3D semantics generation with tradition diffusion losses. 

After generating temporally consistent 3D semantics, we project them onto the image plane using ray casting to obtain four conditions: Semantic Map, Depth Map, Coordinate Map, and Multi-Plane Image (MPI) (see Fig.~\ref{fig:condition}). For each pixel, a ray is cast over the depth range to locate the first non-empty voxel, converting the 3D grid into 2D image representations.
These conditions are grouped based on their roles. The Semantic Map assigns RGB values to semantic labels, the Depth Map normalizes distances, and the Coordinate Map encodes 3D positions into RGB. The MPI, on the other hand, captures semantic details along multiple depth layers. Together, the Semantic and Depth Maps capture surface details, while the MPI and Coordinate Map provide depth and 3D positional context, enhancing geometric fidelity in video generation.

\subsection{3D Geometry-Aware Diffusion Transformer}
\label{sec: video_generation}

As shown in Fig.~\ref{fig:model}, we propose a transformer-based diffusion model and a control encoder for 3D conditions (detailed in Sec.~\ref{sec: guidance_projections}) to generate controllable, 3D-consistent driving videos. The architecture of our transformer and control encoder is described below.

\noindent\textbf{Diffusion Transformer Architecture.}
Our video generation model is built upon the Diffusion Transformer (DiT)-based framework for temporally and multi-view consistent video synthesis. 
Our transformer employs spatial view-inflated attention mechanism that ensures cross-view coherence, 
while introducing significant enhancements tailored to driving scenes. 
The model architecture is illustrated in Figure~\ref{fig:model}(c).
The Control Block extends the Base Block with an Projection layer with zero convolutions.
This stabilizes the initial phase and promotes smoother convergence by minimizing parameter fluctuations.
The generation conditions contain projected 3D bounding boxes, 
layout and 3D semantics-derived guidances, 
as detailed in \secref{sec: guidance_projections}. 
These guidances encode scene geometry and semantics more effectively than 2D conditions.
Text condition is encoded through ~\cite{raffel2020exploring_t5}.

By maintaining inherent geometric consistency through preservation of 3D structural fidelity and cross-depth contextual awareness, 
our framework facilitates accurate cross-view modeling of dynamic interactions. Through systematic integration of these geometrically-constrained projections via our 3D-aware Semantic Guidance Module (\secref{sec: guidance_projections}), 
we establish dense feature correspondences within the ControlNet-enhanced DiT architecture. 
This paradigm significantly improves the model's capacity to synthesize temporally-coherent multi-view sequences with visual fidelity.

\noindent\textbf{3D Guidance Control Encoder.}
To align the feature representations of the 3D semantics-based conditions with those of the original RGB images, 
we adopt two encoding strategies. First, the Semantic, Depth, and Coordinate Maps are processed using the same VAE encoder as the RGB images to ensure a shared latent space. 
Let $\mathbf{I}_v$ denote the RGB image from the $v$-th view and $\mathbf{C}_v^k$ represent the $k$-th condition (with $k \in \{\text{Semantic, Depth, Coordinate}\}$). 
They are encoded as:
\begin{equation}
    \mathbf{z}_{\mathbf{I}_v} = E_{\text{VAE}}(\mathbf{I}_v), \quad \mathbf{z}_{\mathbf{C}_v^k} = E_{\text{VAE}}(\mathbf{C}_v^k),
\end{equation}
where $E_{\text{VAE}}(\cdot)$ maps inputs to a latent space, ensuring alignment for subsequent fusion.

In contrast, the Multi-Plane Image (MPI) encodes depth-aware semantics across multiple planes. 
We design a dedicated MPI encoder that employs lightweight zero convolutions to fuse depth information into a feature representation compatible with the unified latent space. 
This design preserves fine-grained semantic details while aligning MPI features with other 3D semantics conditions.

As described in ~\secref{sec: guidance_projections}, the 3D semantics conditions are divided into two complementary groups, 
each processed by its own encoder. 
Within each group, the encoded features are concatenated along the channel dimension and fused via a zero convolution to produce a unified feature map, which is then fed into the ControlNet (see Fig.~\ref{fig:model}(b)).

\subsection{Multi-Condition Consistency Adapter}
\label{sec: consistency_adapter}

\begin{figure}[ht]
    \centering
    \includegraphics[width=0.85\linewidth]{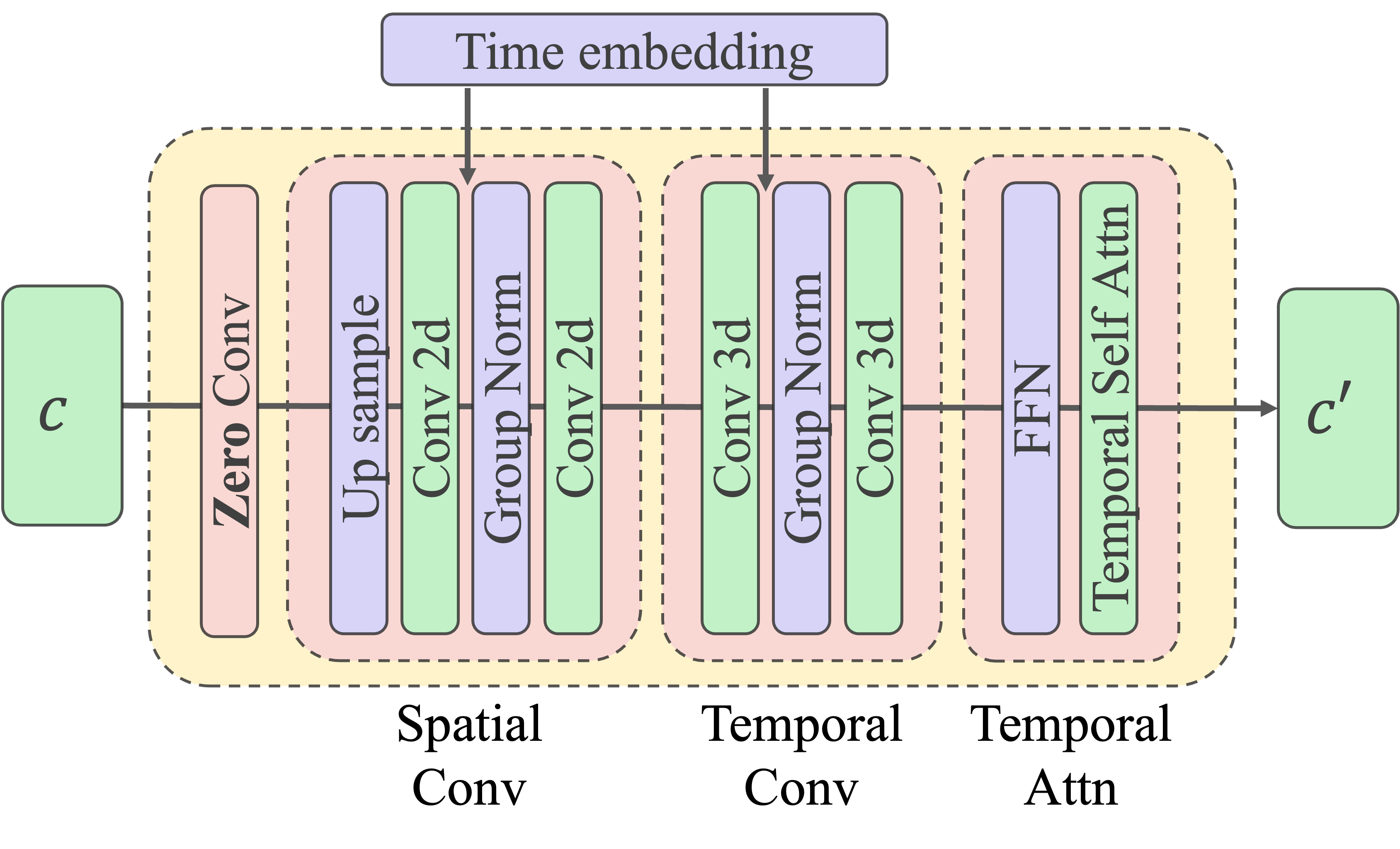}
    \caption{Architecture of the Consistency Adapter. Here, $c$ represents the control conditions output from the control block, and $c'$ denotes the adapter’s output, which replaces $c$ and is integrated into the base block.}
    \label{fig:adapter}
\end{figure}

Training our diffusion transformer with fixed conditions may limit its generative flexibility, hindering adaptation to new conditions such as initial image frames or novel semantic layouts. To address this, we introduce a lightweight Consistency Adapter, 
shown in \figref{fig:adapter}, strategically placed between the output of the control blocks and the base blocks, to enhance the robustness of video generation.

Inspired by the design in~\cite{lin2024ctrladapter}, our Consistency Adapter comprises three sequential modules: a spatial convolution module, a temporal convolution module, and a temporal attention module, each applied once within the adapter. We modify the baseline architecture by removing text cross-attention, avoiding redundancy since scene caption embeddings already perform cross-attention with each block.

To balance efficacy and computational efficiency, we apply the adapter solely to the first two control blocks, where raw semantic and global contextual features predominate. This module is integrated and fine-tuned post-training (as described in Sec.~\ref{sec: video_generation}), ensuring optimal alignment with the pre-trained architecture while reducing computational overhead. Quantitative and qualitative analyses of its impact are presented in our experiments.

\subsection{Training and Inference with Foreground-Mask Loss}
\label{sec: mask_loss}

Our training and inference strategies include rectified flow~\cite{liu2022flow} for stable sampling and first-k frame masking for variable-length generation. 
We further adapt classifier-free guidance~\cite{ho2022classifier} to balance text and multi 3D conditions inputs, boosting controllability.

Moreover, foreground objects (\eg, vehicles, pedestrians) are perceptually critical but cover only a small portion of the image. Standard MSE loss treats all pixels equally, often neglecting fine details in these regions. To mitigate this, we generate binary foreground masks $\mathbf{M} \in \{0,1\}$ via the 3D semantics-guided projection pipeline (\secref{sec: guidance_projections}), where a pixel is marked as foreground ($\mathbf{M}=1$) if its ray intersects an occupied 3D region of interest.
We incorporate the mask into the diffusion loss, yielding:
\begin{equation}
\mathcal{L} = \mathbb{E}\left[\|e\|_2^2 + \gamma\,\|\mathbf{M}\odot e\|_2^2\right], \quad e = \epsilon - \epsilon_\theta(z_t,t,c),
\end{equation}
where $\epsilon$ is the ground-truth noise and $\epsilon_\theta(z_t,t,c)$ is the predicted noise. This formulation enhances foreground details while preserving overall video quality.

\section{Experiments}

\begin{figure*}[ht]
    \centering
    \includegraphics[width=0.95\linewidth]{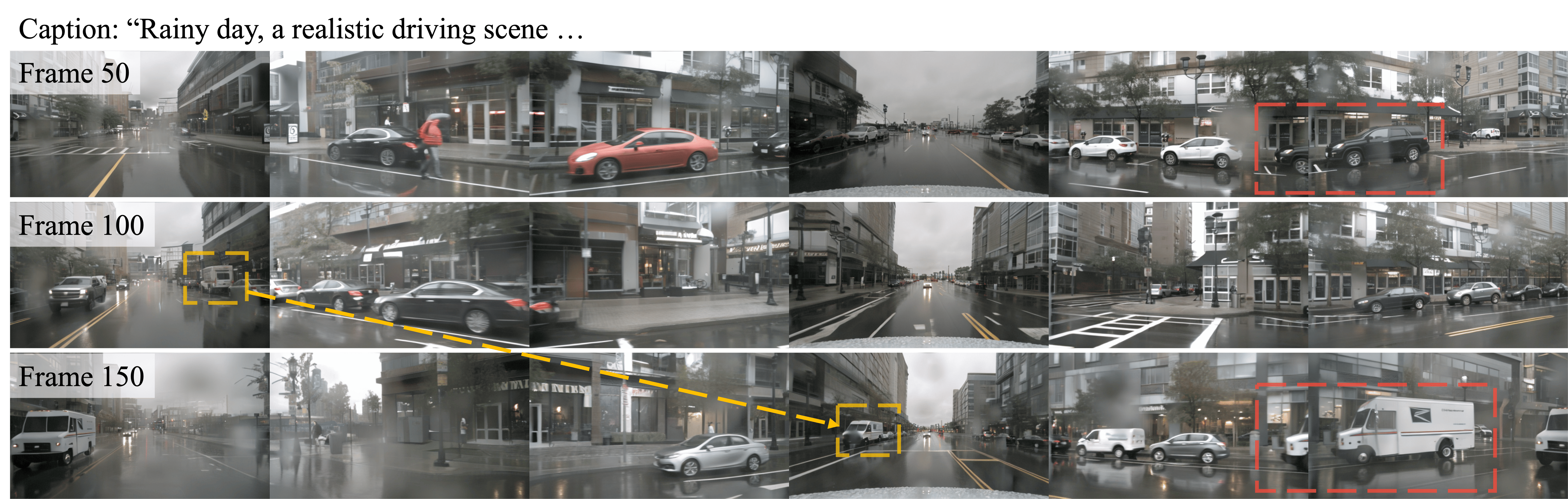}
    \caption{An example of out generated long driving video. Contents in red boxes indicate that adjacent viewpoints maintain the same appearance.The yellow arrows show that the generated images remain consistent over different time steps.} 
    \label{fig:gen_vis}
\end{figure*}

\begin{figure*}[ht]
    \centering
    \includegraphics[width=0.95\linewidth]{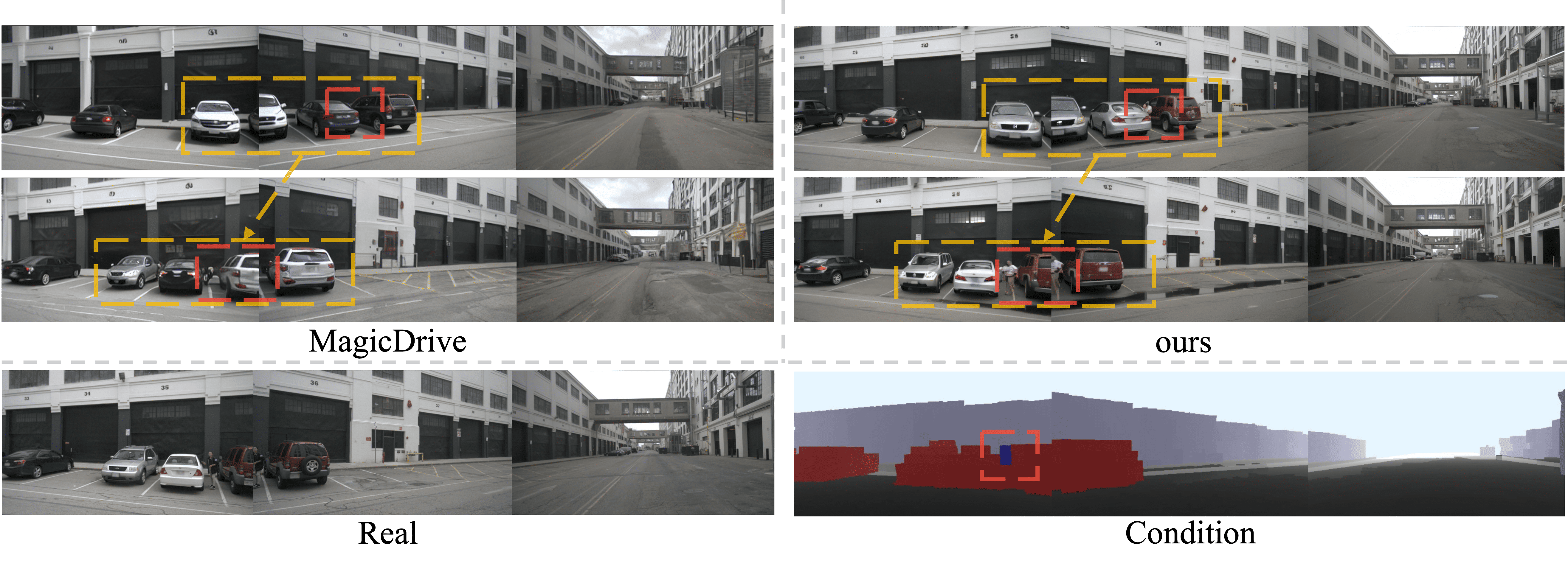}
    \caption{Qualitative comparison with MagicDrive. Our approach exhibits enhanced spatial-temporal consistency and finer details, particularly noticeable in the foreground objects and scene geometry.} 
    \label{fig:magicdrive}
\end{figure*}

\subsection{Experimental Setups}
\label{sec:experimental_setups}

\paragraph{Dataset and Baselines.}
We conduct our experiments on the publicly available nuScenes dataset~\cite{caesar2020nuscenes}, a large-scale autonomous driving benchmark containing 1000 driving scenes in urban environments. Each scene covers approximately 20 seconds and provides annotations at 2Hz. To ensure continuous and sufficient data, we interpolate the 2Hz annotations to 12Hz~\cite{wang2023we_asap}. We train our model on the 700-scene training set and evaluate it on the 150-scene validation set.
We adopt DiVE~\cite{jiang2024dive} as our baseline method. It integrates a parameter-free spatial view-inflated attention module to maintain cross-view coherence and uses road layouts and bounding boxes as conditions. Comparisons against DiVE highlight our 3D semantics-centric approach's effectiveness in preserving visual fidelity and geometric consistency across different views.

\paragraph{Evaluation Metrics.}
Following \cite{du2024challenge}, we evaluate our method from two perspectives: \emph{generation quality} and \emph{controllability}. For quality assessment, we employ the Fréchet Video Distance (FVD)~\cite{unterthiner2019fvd}, a widely adopted metric to assess how closely the distribution of generated videos matches that of real-world data, and the Fréchet Inception Distance (FID), which measures the similarity between generated and real images at the feature level. 
To measure controllability, we focus on two prominent tasks in autonomous driving: 3D object detection and BEV segmentation. We follow the nuScenes official protocol to compute mean Average Precision (mAP) for 3D detection and use mean Intersection over Union (mIoU) for BEV segmentation. Both tasks are conducted using the video-based perception model BEVFormer~\cite{li2024bevformer}. 
We retain the first 16 frames for evaluation, consistent with \cite{du2024challenge,gao2024magicdrivedit}.

\paragraph{Model Setup.}
We select the pretrained OpenSora-VAE-1.2~\cite{OpenSora-VAE-v1.2} with frozen parameters, exclusively training the Diffusion Transformer. 
The training protocol comprises two phases: 
During the initial multi-resolution pretraining, we progressively scale input resolutions from 144p ($144\times256$) to 900p ($900\times1600$), 
processing video clips of varying lengths—longer sequences up to 128 frames at lower resolutions and shorter sequences down to 6 frames at higher resolutions. 
This is followed by an adapter fine-tuning phase where lightweight spatiotemporal adapters are inserted into the DiT blocks, 
with fixed configurations at 360p ($360\times640$) and 16 frames. 
For inference, we select 360p resolution, 
with the first frame ground truth observations as the reference frame, which is aligned with prior works~\cite{gao2024magicdrivedit,li2024uniscene}.

\begin{table}[ht]
    \centering
    \small
    \begin{tabular}{l|c|c|cc}
        \toprule
        Method        & FPS & Resolution  & FVD$\downarrow$ & FID$\downarrow$ \\
        \midrule
        MagicDrive~\cite{gao2023magicdrive}            & 12Hz& 224$\times$400 & 218.12         & 16.20           \\
        Panacea~\cite{wen2024panacea}                  &  2Hz & 256$\times$512& 139.00         & 16.96           \\
        SubjectDrive~\cite{huang2024subjectdrive}      &  2Hz & 256$\times$512 & 124.00         & 15.98           \\
        DriveWM~\cite{wang2024driving_drivewm}         &  2Hz & 192$\times$384 & 122.70         & 15.80           \\
        Delphi~\cite{ma2024delphi}                     & 2Hz &  512$\times$512 & 113.50         & 15.08           \\
        MagicDriveDiT~\cite{gao2024magicdrivedit}      & 12Hz & 224$\times$400 & 94.84          & 20.91           \\
        DiVE~\cite{jiang2024dive}                      & 12Hz & 480$\times$854 & 94.60          & -               \\
        \midrule
        Ours & 12Hz & 360$\times$640 & \textbf{68.43} & \textbf{10.15} \\

        \bottomrule
    \end{tabular}
    \caption{Quantitative comparison on video generation quality with other methods. Our method achieves the best FVD score.}
    \label{tab:fvd_fid}
\end{table}

\begin{table}[ht]
    \centering
    \setlength{\tabcolsep}{6pt}
    \begin{tabular}{l|cc}
        \toprule
        Method        & mIoU$\uparrow$ & mAP$\uparrow$ \\
        \midrule
        MagicDrive~\cite{gao2023magicdrive}  & 18.34          & 11.86         \\
        MagicDrive3D~\cite{gao2024magicdrive3d}  & 18.27          & 12.05         \\
        MagicDriveDiT~\cite{gao2024magicdrivedit} & 20.40          & 18.17         \\
        DiVE~\cite{jiang2024dive}          & 35.96          & 24.55         \\
        \midrule
        Ours          &    \textbf{37.80}           &      \textbf{27.88}    \\
        \bottomrule
    \end{tabular}
    \caption{Comparison with baselines for video generation controllability. Results are calculated with first 16 frames of videos.}
    \label{tab:miou_map}
\end{table}

\begin{table}[ht]
    \centering
    \small
    \setlength{\tabcolsep}{3pt}
    \begin{tabular}{c|cc|cc|c|c|cc}
        \toprule
        Index & Sem & Dep & MPI & Coor & Adapter & 3D-Sem & FVD$\downarrow$ & FID$\downarrow$ \\ 
        \midrule
        (0)& & & & & & & 105.56 & 19.46 \\
         (1)& &  & \checkmark & \checkmark & \checkmark & GT & 71.10 & 11.49 \\
         (2)& &  & \checkmark & \checkmark & \checkmark & GEN & 72.04 & 11.70 \\
         \midrule
        (3)& \checkmark & & & & & GT & 72.67 & 11.73 \\
        (4)& \checkmark & \checkmark & & & & GT & 69.54 & 11.39 \\
         (5)& \checkmark & \checkmark &  &  & \checkmark & GT & \textbf{68.43} & \textbf{10.15} \\
        (6)& \checkmark & \checkmark & & & \checkmark & GEN & 70.85 & 11.38 \\
        \bottomrule
    \end{tabular}
    \caption{Ablation results on 16-frame video generation. `Sem Dep' denotes Semantic Map and Depth Map, while `MPI Coor' refers to MPI and Coordinate Map.
        `Adapter' indicates the consistency adapter, and `3D-Sem' represents 3D semantics-based guidance (GT for ground truth, GEN for our generated 3D semantics).}
    \label{tab:ablation}
\end{table}

\subsection{Results and Analysis}
\label{sec:results}

\paragraph{Generation Quality.}
An example of our generated driving scene video is shown in Figure~\ref{fig:gen_vis}. 
The video captures rich details with high realism while maintaining strong consistency across different views, 
ensuring seamless transitions between adjacent camera perspectives. 
Each displayed frame is sampled approximately 4.2 seconds apart, demonstrating stable temporal coherence and smooth motion continuity throughout the video, up to the final frame.
Figure~\ref{fig:magicdrive} illustrates a qualitative comparison with MagicDrive, showcasing our method's enhanced realism.

We conduct comprehensive evaluations of video generation quality, as shown in Table~\ref{tab:fvd_fid}. 
In our method, the 3D semantics guidance is derived from ground truth using Semantic Maps and Depth Maps (S + D). 
Our approach achieves state-of-the-art performance, attaining an FVD of 68.43, which outperforms traditional layout/bbox-guided methods~\cite{jiang2024dive,gao2024magicdrivedit,ma2024delphi} and 3D semantics-guided methods~\cite{li2024uniscene}. 
The superior FVD indicates enhanced motion consistency, while the competitive FID confirms that visual quality is well preserved. 
These results demonstrate that utilizing ground truth 3D semantics guidance can effectively boost video generation performance.

\paragraph{Downstream Task Utility}
The effectiveness of our method in enhancing controllability for downstream perception tasks is demonstrated in Table~\ref{tab:miou_map}. 
Our method significantly outperforms baseline approaches, achieving an mIoU of 37.80 and an mAP of 27.88. 
Specifically, it surpasses DiVE by 5.1\% in mIoU and 13.6\% in mAP, and MagicDriveDiT by 85.3\% in mIoU and 53.4\% in mAP. 
These improvements highlight that our 3D semantics-based representation of foreground object shapes, positions, and map elements offers superior controllability compared to traditional bounding boxes and semantic layout conditions. 
Such advancements underscore the practical utility of our approach for real-world applications, where precise spatial understanding directly impacts downstream task performance.

\subsection{Ablation Studies}
\label{sec:ablation_studies}

\begin{figure*}[ht]
    \centering
        \centering
        \includegraphics[width=0.95\linewidth]{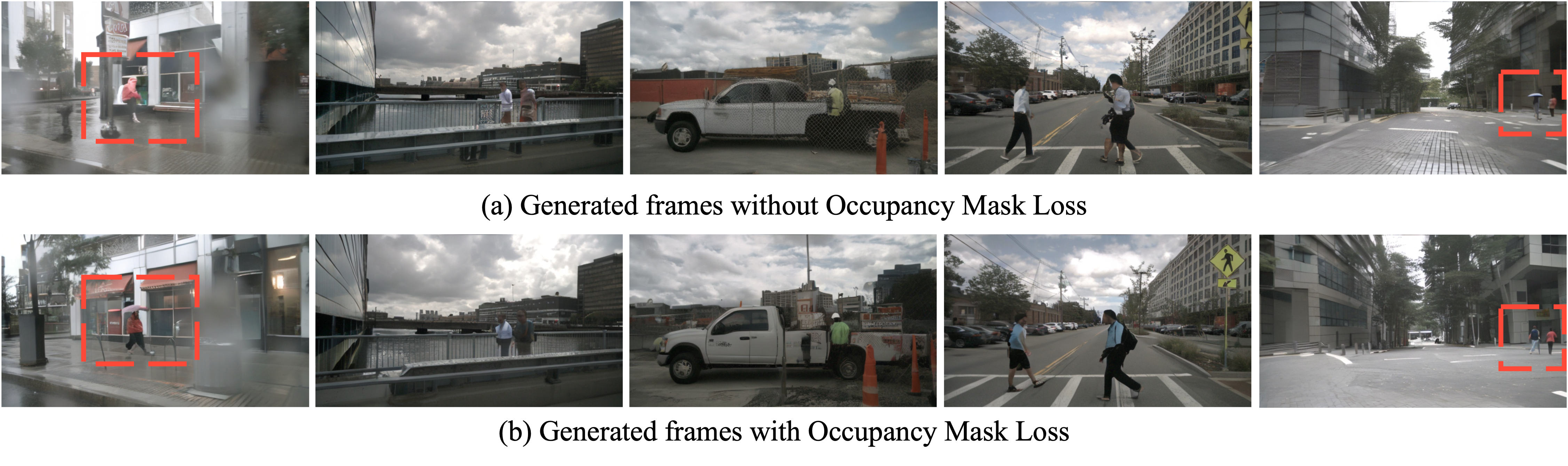}
        \label{fig:wo_mask_loss}
    \caption{Qualitative visualization of 3D semantics Mask Loss's effects. The top panel (without 3D semantics Mask Loss) finds blurred edges and missing details in foreground pedestrians, particularly for distant targets. The bottom panel (with 3D semantics Mask Loss) shows sharper outlines and enhanced textures in foreground pedestrians, especially in distant regions, highlighting the effectiveness of the mask loss in improving detail fidelity.}
    \label{fig:mask_loss}
\end{figure*}

\begin{figure}[ht]
    \centering
    \includegraphics[width=0.95\linewidth]{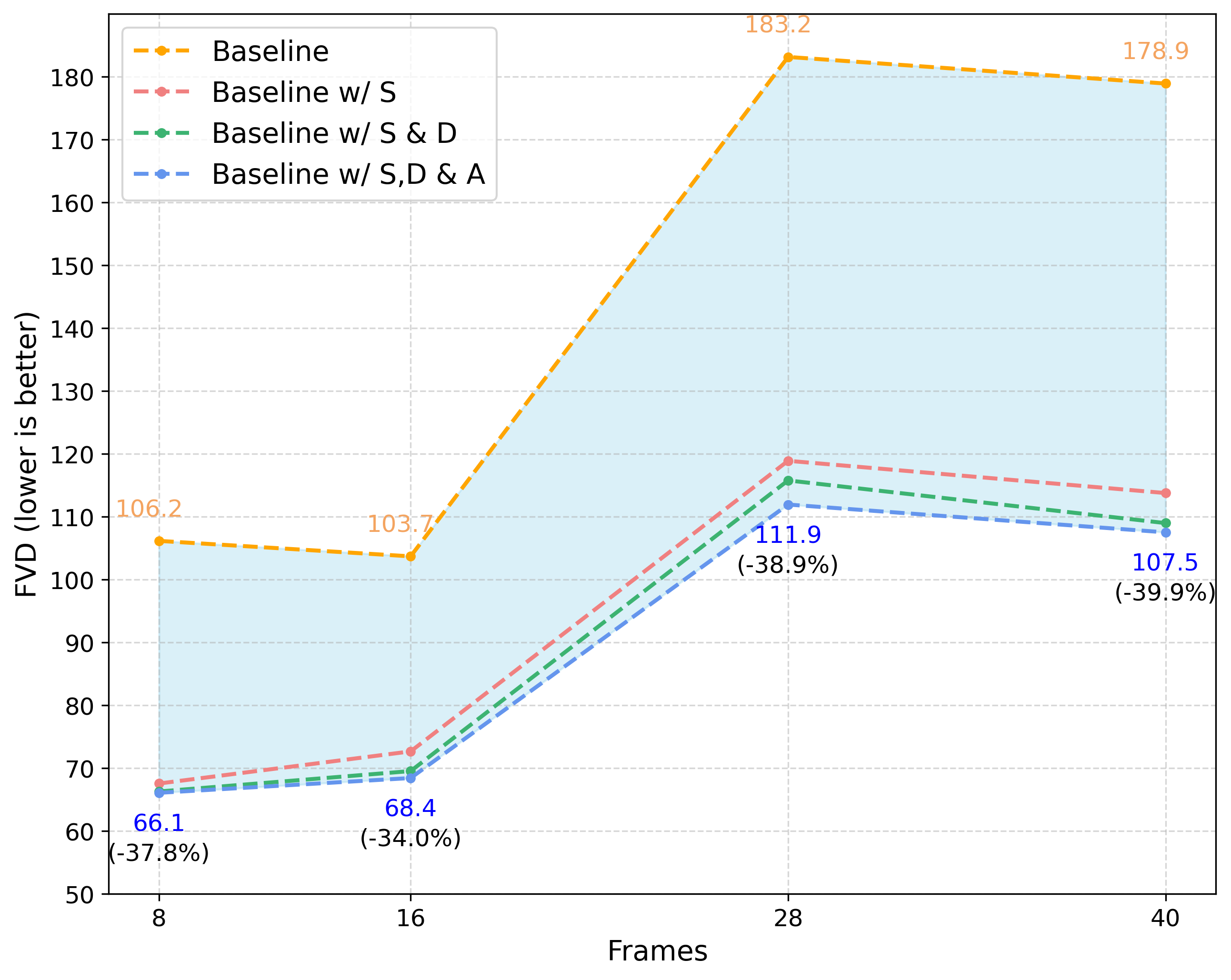}
    \caption{FVD scores comparison for different model settings. Incorporating the adapter consistently lowers FVD across 8, 16, 28, and 40 frame sequences.} 
    \label{fig:fvd_comparison_all}
\end{figure}

\paragraph{Studies on 3D semantics Conditions}
Table~\ref{tab:ablation} presents the ablation results for various 3D semantics guidance configurations. 
Group (3) shows that using only the Semantic Map yields an FVD of 72.67 and an FID of 11.73, surpassing the baseline by a large margin. 
In Group (4), incorporating the Depth Map together with the Semantic Map improves performance to an FVD of 69.54 and an FID of 11.39, demonstrating that the addition of depth information introduces essential spatial details. 
Group (5) indicates that adding the consistency adapter to the combination of the Semantic Map and the Depth Map reduces the FVD to 68.43 and the FID to 10.15, signifying a notable enhancement in temporal consistency. 
Group (1), which uses MPI and Coordinate Maps together with the consistency adapter, achieves an FVD of 71.10 and an FID of 11.49. 
As seen in Group (2) versus Group (1) and Group (6) versus Group (5), comparisons between GT and GEN consistently show that guidance based on ground truth yields better performance, confirming the advantage of high-fidelity 3D semantics information.
More results can be found in Figure~\ref{fig:fvd_comparison_all}.

\paragraph{Studies on Consistency Adapter}

We compared our model with and without the adapter across different sequence lengths: 8, 16, 28, and 40 frames. Quantitative results are shown in Table~\ref{tab:ablation}, where adding the consistency adapter reduces the FVD by 1.11. As illustrated in Figure~\ref{fig:fvd_comparison_all}, the model with the adapter (blue line) consistently shows lower values than the model without the adapter (green line). These improvements underscore the importance of incorporating temporal smoothing mechanisms in video generation tasks. This highlights the adapter’s efficiency in maintaining spatiotemporal consistency in extended video sequences, optimizing inter-frame relationships, and mitigating jitter.

\paragraph{Studies on Mask Loss}

To assess the impact of the proposed 3D semantics Mask Loss, we trained our models with different strategies: one optimized with the normal MSE loss and another incorporating the 3D semantics Mask Loss, as described in Section~\ref{sec: mask_loss}. The qualitative results, presented in Figure~\ref{fig:mask_loss}, demonstrate significant improvements in foreground object detail, particularly for pedestrians. The top panel of the figure (training with only MSE loss) exhibits blurred edges and fewer details in foreground pedestrians, especially for distant targets. In contrast, with 3D Semantics Mask Loss, frames in the bottom panel show sharper outlines and enhanced textures in foreground pedestrians, particularly in distant regions, highlighting the mask loss’s effectiveness in enhancing detail fidelity. These findings suggest that integrating the 3D semantics Mask Loss is critical for preserving fine-grained details in complex scenes.

\paragraph{Partial 3D Semantics Usage for Training}

Acquiring high-quality 3D semantics information for every clip can be prohibitively expensive in real-world scenarios. 
Therefore, we explore a hybrid training strategy in which only 50\% of the training samples utilize 3D semantics as the primary guidance, while the remaining 50\% rely on traditional layout and bounding-box conditions (BEV). 
Table~\ref{tab:bev_occ} shows that the partially 3D semantics-trained model (50\%/50\%) achieves FVD scores of 73.82 and 73.80 when tested with 3D Sem and BEV guidance, respectively—only a moderate degradation compared to the fully 3D semantics-trained counterpart (FVD 68.43). This demonstrates that the model maintains a robust generative performance even with partial 3D semantics guidance.
Moreover, this adaptive capacity highlights the model's flexibility in handling varying guidance signals, which is essential for real-world applications where data quality can be inconsistent.

\begin{table}[ht]
    \centering
    \small
    \setlength{\tabcolsep}{3pt}
    \begin{tabular}{cc|cc}
        \toprule
        Train (3D-SEM/BEV) & Infer (3D-SEM/BEV) & FVD$\downarrow$ & FID$\downarrow$ \\
        \midrule
         0\% / 100\% & 100\% / 0\% & 120.65 & 20.45 \\
         0\% / 100\% & 0\% / 100\% & 103.70 & 19.46 \\
         100\% / 0\% & 100\% / 0\% & 68.43  & 10.15 \\
        \midrule
         50\% / 50\% & 100\% / 0\% & 73.82  & 11.74 \\
         50\% / 50\% & 0\% / 100\% & 73.80  & 11.86 \\
        \bottomrule
    \end{tabular}
    \caption{Comparative results for training and inference under varying 3D Semantics and traditional layout/bounding box (BEV) guidance ratios. 
    Each cell shows ``3D-SEM\% / BEV\%''. Performance is evaluated on the first 16 frames.}
    \label{tab:bev_occ}
\end{table}

\section{Conclusion}

In this paper, we introduced CoGen, a novel framework designed to generate high-quality driving videos with enhanced photorealism and 3D consistency, 
leveraging detailed 3D semantics information. 
By incorporating semantics guidance in multiple forms, foreground-aware mask loss training, and a consistency adapter module, 
CoGen further improves video quality and 3D coherence.
Experimental results on the nuScenes dataset demonstrate state-of-the-art performance with an FVD of 68.43, 
outperforming existing methods based on 2D layout and semantics guidance. 
Moreover, our generated videos exhibit remarkable utility in downstream perception tasks. 
These results demonstrate the practical value of our method in generating synthetic data for autonomous driving, preserving geometric fidelity and visual realism.

{
    \small
    \bibliographystyle{ieeenat_fullname}
    \bibliography{cite}
}
\clearpage

\end{document}